\documentclass[letterpaper, 10 pt, conference]{ieeeconf}  

\IEEEoverridecommandlockouts                              

\usepackage{graphicx}      
\usepackage{algorithm} 
\usepackage{algpseudocode} 
\usepackage{amsfonts}
\usepackage{amsmath}
\usepackage{url}
\usepackage{bbm} 
\usepackage{tabularx}
\usepackage[font=small,labelfont=bf]{caption}

\usepackage{subcaption}
\usepackage{upgreek}

\newcommand{\argmin}{\mathop{\mathrm{argmin}}}  
\newcommand{\argmax}{\mathop{\mathrm{argmax}}}  
\graphicspath{{figures/}}
\title{\LARGE \bf
A Learning Framework for High Precision Industrial Assembly
}
\author{Yongxiang Fan$^{1}$, Jieliang Luo$^{2}$, Masayoshi Tomizuka$^{1}$
\thanks{Yongxiang Fan and Masayoshi Tomizuka are with 
	University of California, Berkeley, Berkeley, CA 94720, USA
	{\tt\small {yongxiang\_fan, tomizuka}@berkeley.edu}}%
\thanks{Jieliang Luo is with 
	University of California, Santa Barbara,  Santa Barbara, CA 93106, USA 
	{\tt\small jieliang@ucsb.edu}}%
}

\begin{document}
\maketitle
\thispagestyle{empty}
\pagestyle{empty}

\begin{abstract}
Automatic assembly has broad applications in industries. Traditional assembly tasks utilize predefined trajectories or tuned force control parameters, which make the automatic assembly time-consuming, difficult to generalize, and not robust to uncertainties. In this paper, we propose a learning framework for high precision industrial assembly. The framework combines both the supervised learning and the reinforcement learning. The supervised learning utilizes trajectory optimization to provide the initial guidance to the policy, while the reinforcement learning utilizes actor-critic algorithm to establish the evaluation system even the supervisor is not accurate. The proposed learning framework is more efficient compared with the reinforcement learning and achieves better stability performance than the supervised learning. 
The effectiveness of the method is verified by both the simulation and experiment. Experimental videos are available at~\cite{website}.
\end{abstract}

\section{Introduction}
Automatic precision assembly is important for industrial manipulators to improve the efficiency and reduce the cost. Most of the current assembly tasks rely on dedicated manual tuning to provide trajectories for specific tasks, which requires intensive labors and is not robust to uncertainties. To reduce the human involvement and increase the robustness to uncertainties, more researches are focusing on learning the assembly skills. 

There are three types of learning in Psychology~\cite{kalat2016introduction}: classical conditioning, observational learning and operant conditioning. The second and third types correspond to supervised learning and reinforcement learning, respectively. 
The supervised learning is ideal when the training data is sufficient. Practically, collecting data is inefficient under various uncertainties of the environment. A Gaussian mixture model (GMM) is trained in~\cite{tang2016teach} from human demonstration to learn a peg hole insertion skill. The peg hole insertion task is simplified by constraining the policy into planar motion and the trained policy is not adaptable to different environments. 

The reinforcement learning (RL) learns a sequence of optimal actions by exploring the environment to maximize the expected reward. Different types of RL methods include the direct policy gradient such as REINFORCE~\cite{williams1992simple}, Q-learning based methods such as DQN~\cite{mnih2015human}, as well as the actor-critic framework such as  DDPG~\cite{lillicrap2015continuous} or PPO~\cite{schulman2017proximal}. 
These methods are called model-free RL since the dynamics model is not used during exploration. Despite lack of dynamics, the model-free RL has been successfully applied to assembly tasks~\cite{vecerik2017leveraging, inoue2017deep}. The model-free RL requires considerable data to explore the state/action space and reconstruct the transitions of the environment. Consequently, it is less data-efficient and time-efficient. 

Model-based RL is proposed to increase the data efficiency~\cite{levine2013guided, levine2016end}. It fits dynamics models and applies optimal control such as iLQR/iLQG~\cite{tassa2012synthesis} to compute the optimal trajectories. The exploration is conducted by adding random noise to the actions during the optimization. Then the optimized trajectories are used to train a neural network policy in a supervised manner. Compared with model-free RL, the model-based RL has larger exploit-exploration ratio, thus explores narrower space and converges faster than the model-free RL. The performance of the model-based RL depends on the behavior of the optimal controller (i.e. supervisor), which in turn is effected by the accuracy of the local dynamics model. For the rigid robot dynamics with force/torque as states, the dynamics model is less smooth\footnote{The dynamics change dramatically as the trajectory slightly changes.}, which makes the dynamics fitting not effective. Consequently, the model-based RL cannot converge consistently. In practice, people usually use soft robotics model (Baxter, PR2)~\cite{levine2016end} with position/velocity states by ignoring the force/torque feedback. 

This paper proposes a learning framework to train a more natural assembly policy by incorporating both the force/torque and the positional feedback signals. The proposed framework combines the model-based RL with the model-free actor-critic to learn the manipulation skills for precision assembly tasks. 
The model-based RL computes for the optimal trajectories with both positional and force/torque feedback. The performance of the controller might be affected by the smoothness of the local fitted dynamics model. To avoid the problem of inconsistency or tedious parameter tuning of optimal controller, a critic network is introduced to learn the correct critic value (Q-value). Instead of training the policy network by pure supervision, we train an actor network by combining the supervised learning with the policy gradient. To accelerate the training efficiency of the critic network, the Q-value from the optimal control is employed to train the critic network. 

The contribution of this work are as follows. First, the optimal controller is able to constrain the exploration space in safe region compared with the random exploration at the first iterations of actor-critic methods. 
Secondly, the optimal controller is more data-efficient when exploring in a narrower space and solving for optimal trajectory mathematically. 
Thirdly, the combined critic network is able to address the potential inconsistency and instability of the optimal controller caused by the rigid robotics system and force/torque feedback, and build up a ground truth critic for the policy network. 

The remainder of this paper is described as follows. The related work is stated in Section~\ref{sec:related}, followed by a detailed explanation of the proposed learning framework in Section~\ref{sec:proposed}. Simulation and experiment results are presented in Section~\ref{sec:experiments}. Section~\ref{sec:conclusions} concludes the paper and proposes future works.

\section{Related Work}
\label{sec:related}
The objective of an assembly task is to learn an optimal policy $\pi_\theta(a_t|o_t)$ to choose an action $a_t$ based on the current observation $o_t$ in order to maximize an expected reward: 
\begin{equation} 
\label{eq:loss_minimization}
\begin{aligned}
\min_{\pi_\theta} E_{\tau \sim \pi_\theta}(l(\tau)),
\end{aligned}
\end{equation}
where $\theta$ is the parameterization of the policy, $\tau = \{s_0,a_0, s_1, a_1, ..., s_T, a_T\}$ is the trajectory, $\pi_\theta(\tau) = p(s_0)\prod_{1}^{T}p(s_t|s_{t-1},a_{t-1})\pi_\theta(a_t|s_t)$, and $l$ is the loss of the trajectory $\tau$. 

Equation~(\ref{eq:loss_minimization}) can be solved by optimization once a global dynamics model $p(x_t|x_{t-1}, u_{t-1})$ is explicitly modeled. For a contact-rich complex manipulation task, the global dynamics model is extremely difficult to obtain. Therefore, the assembly task either avoids using dynamics~\cite{inoue2017deep} or fits the a linear dynamics model~\cite{tang2016teach,levine2013guided, levine2016end}. 

On one hand, the RL without dynamics requires excessively data to explore the space and locate to the optimal policy due to the potential high-dimensionality of the action space. On the other hand, the performance of the~\cite{levine2013guided, levine2016end} can be downgraded once the robotic system is rigid or the force/torque feedback is included in the optimal controller. 

We propose a learning framework that combines the actor-critic framework and optimal control for efficient high-accuracy assembly. The optimal controller is adapted from the model-based RL~\cite{levine2013guided}, while the actor-critic framework is modified from the DDPG algorithm. 
These two algorithms will be briefly introduced below.

\subsection{Deep Deterministic Policy Gradient (DDPG)}

The DDPG algorithm collects sample data $(s_j, a_j, s_{j+1}, r_j)$ from the replay buffer $R$ and trains a critic network ${Q}_\phi$ and actor network $u_\theta$ parameterized by $\phi$ and $\theta$. 
More specifically, the critic network is updated by:
\begin{equation} 
\label{eq:ddpg_critic}
\begin{aligned}
& \phi \leftarrow \argmin_\phi\frac{1}{N_{dd}}\sum_{j = 1}^{N_{dd}}\left(y_j - {Q}_{{\phi}}(s_j, a_j)\right)^2, \\
& y_j = r_j + \gamma {Q}_{\hat{\phi}}(s_{j+1}, u_{\hat{\theta}}(s_{j+1})),
\end{aligned}
\end{equation}
where $N_{dd}$ is the batch size for DDPG, $\hat{\phi}, \hat{\theta}$ are parameters of the target critic network and target actor network, and $\gamma$ is the discount for future reward.  

The policy network is updated by:
\begin{equation} 
\label{eq:ddpg_actor}
\begin{aligned}
\theta \leftarrow \argmax_\theta\frac{1}{N_{dd}}\sum_{j = 1}^{N_{dd}} {Q}_{\hat{\phi}}(s_j, u_\theta(s_j)),
\end{aligned}
\end{equation}
where $\theta$ is the parameters for the policy network to be optimized. Policy gradient is applied to update the parameters of the actor network: 
\begin{equation} 
\label{eq:ddpg_policy_gradient}
\begin{aligned}
\theta \leftarrow \theta + \alpha \frac{1}{N_{dd}}\sum_{j = 1}^{N_{dd}}\nabla_a \hat{Q}(s,a)|_{s = s_j, a=a_j} \nabla_\theta u_\theta(s)|_{s = s_j},
\end{aligned}
\end{equation}
where the $\alpha$ is the learning rate of the actor network. 

The target networks are updated by
\begin{equation} 
\label{eq:ddpg_target_update}
\begin{aligned}
&\hat{\phi} \leftarrow \delta \phi + (1 - \delta) \hat{\phi}, \\
&\hat{\theta} \leftarrow \delta \theta + (1 - \delta) \hat{\theta},
\end{aligned}
\end{equation}
where $\delta$ is the target update rate and is set to be small value ($\delta\approx 0.01$).

\subsection{Guided Policy Search (GPS)}
With the involvement of guiding distribution $p(\tau)$,Problem~(\ref{eq:loss_minimization}) can be rewritten as 
\begin{equation} 
\label{eq:gps_minimization}
\begin{aligned}
\min_{\pi_\theta, p} E_{p}(l(\tau)), \quad s.t. \ \  p(\tau) = \pi_\theta(\tau).
\end{aligned}
\end{equation}

GPS solves~(\ref{eq:gps_minimization}) by alternatively minimizing the augmented Lagrangian with respect to primal variables $p, \pi_\theta$ and updating the Lagrangian multipliers~$\lambda$.  The augmented Lagrangian for $\theta$ and $p$ optimization are:  
\begin{equation} 
\label{eq:gps_Lagrangian}
\begin{aligned}L_p(p, \theta)  = & E_p(l(\tau)) + \lambda \left(\pi_\theta(\tau)- p(\tau)\right) + \\ 
& \nu D_{KL}\left(p(\tau)\|\pi_\theta(\tau)\right),\\
L_\theta(p, \theta)  = & E_p(l(\tau)) + \lambda \left(\pi_\theta(\tau)- p(\tau)\right) + \\ 
& \nu D_{KL}\left(\pi_\theta(\tau)\|p(\tau)\right),
\end{aligned}
\end{equation}
where $\lambda$ is the Lagrangian multiplier, $\nu$ is the penalty parameter for the violation of the equality constraint, and $D_{KL}$ represents the KL-divergence. 
The optimization of primal variable $p$ is called trajectory optimization. It optimizes the guiding distribution $p$ with learned local dynamics. To assure the accuracy of dynamics fitting, the optimization is constrained within the trust region $\epsilon$: 
\begin{equation} 
\label{eq:gps_trajopt}
\begin{aligned}
\min_{p} L_p(p, \theta), \quad s.t. \ \  D_{KL}(p(\tau)\|\hat{p}(\tau)) \leq \epsilon,
\end{aligned}
\end{equation}
where $\hat{p}$ is the guiding distribution of the previous iteration.  The Lagrangian of~(\ref{eq:gps_trajopt}) is:
\begin{equation} 
\label{eq:gps_trajopt_lagrangian}
\begin{aligned}
\mathcal{L}(p) = L_p(p,\theta) + \eta(D_{KL}(p(\tau)\|\hat{p}(\tau)) - \epsilon),
\end{aligned}
\end{equation}
where $\eta$ is the Lagrangian multiplier for the constraint optimization. With the Gaussian assumption of the dynamics,~(\ref{eq:gps_trajopt_lagrangian}) is solved by iLQG. To avoid large derivation from the fitted dynamics, $\eta$ is adapted by comparing the predicted KL-divergence with the actual one. 

The optimization of the policy parameters $\theta$ can be written as a supervised learning problem. With the Gaussian policy $\pi_\theta(a_t|o_t) = \mathcal{N}(u_\theta(o_t), \Sigma^\pi_t)$, we can rewrite $L_\theta(p,\theta)$ in~(\ref{eq:gps_Lagrangian}) as: 
\begin{equation} 
\label{eq:gps_polopt_objective}
\begin{aligned}
&L_\theta(\theta, p) =  \frac{1}{2N_b}\sum_{i,t=1}^{N_b,T}E_{p_i(s_t,o_t)}[
\text{tr}\left(C_{ti}^{-1}\Sigma^\pi_t
\right)-\text{log}|\Sigma^\pi_t|+\\&
\left(u_\theta(o_t) - u_{ti}^p(s_t)\right)^TC_{ti}^{-1}\left(u_\theta(o_t) - u_{ti}^p(s_t)\right) + 2\lambda^T_{t}u_\theta(o_t)
],
\end{aligned}
\end{equation}
where $p_i(u_t|s_t)\sim \mathcal{N}(u_{ti}^p(s_t), C_{ti})$ is the guiding distribution. 
Equation~(\ref{eq:gps_polopt_objective}) contains the decoupled form of the variance optimization and policy optimization. 
Refer~\cite{levine2016end} for more details.

\begin{figure}[t]
	\begin{center}
		{\includegraphics[width =\linewidth]{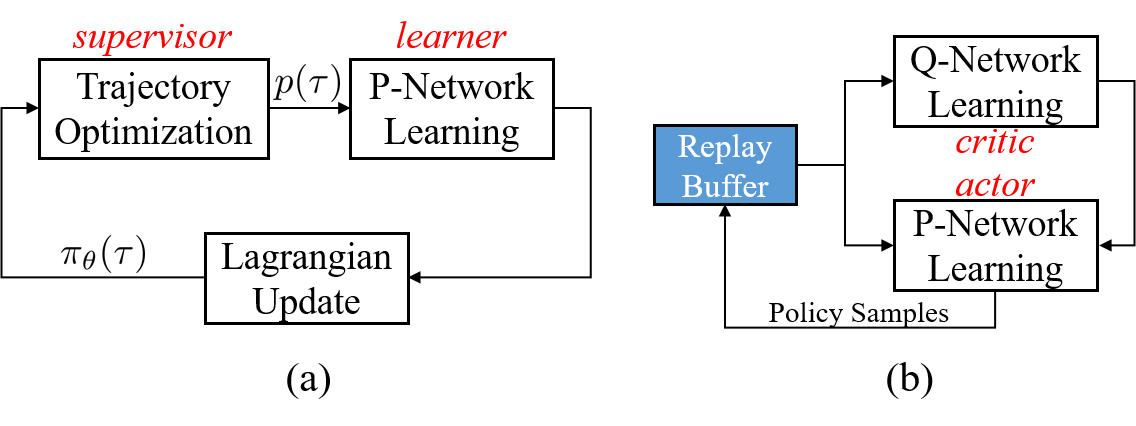}}
		\caption{(a) Guided Policy Search (GPS) and (b) Deep deterministic policy gradient (DDPG). }
		\label{fig:GPS-DDPG}
	\end{center}
\end{figure}
\subsection{Comparison of GPS and DDPG}
GPS decouples RL into a trajectory optimization (\textit{supervisor}) and a supervised policy network learning (\textit{learner}), as shown in Fig.~\ref{fig:GPS-DDPG}(a). The performance of the learner relies on the quality of the supervisor. By fitting the dynamics from sampling data and computing the supervisor with the optimal control, GPS is more efficient than the DDPG and many other model-free RL algorithms. 
However, the performance of the learner would be compromised if the system has high stiffness and has force/torque feedback as states due to the less smooth dynamics and smaller trust region. 

In comparison, DDPG uses rollout samples to jointly train the Q-network (\textit{critic}) and policy network (\textit{actor}), as shown in Fig.~\ref{fig:GPS-DDPG}(b). The critic gradually builds up the Q-value from physical rollouts, and the Q-value is applied to train the actor network based on policy gradient. 
The actor-critic framework provides more stable policy in the tasks with non-smooth dynamics. These tasks are common in high precision industrial assembly where the system has higher stiffness and contains force/torque feedback in the states. Despite the reliable performance, the actor-critic framework is less data efficient due to the intensive exploration, which is usually unnecessary since assembly tasks only requires exploration in narrow trajectory space.

\section{Proposed Approach}
\label{sec:proposed}
Precision industrial assembly usually has large system stiffness in order to achieve precise tracking performance and reduce the vibration.  
With large stiffness, small clearance and force/torque feedback, both the model-free RL and model-based method cannot accomplish the task efficiently and stably. 
In this paper, we propose a learning framework that combines the actor-critic with the model-based RL for high precision industrial assembly. The framework is named as guided-deep deterministic policy gradient (guided-DDPG). Guided-DDPG behaves more efficient than the actor-critic and more stable/reliable than the model-based RL. 

 \begin{figure}[t]
	\begin{center}
		{\includegraphics[width =0.8\linewidth]{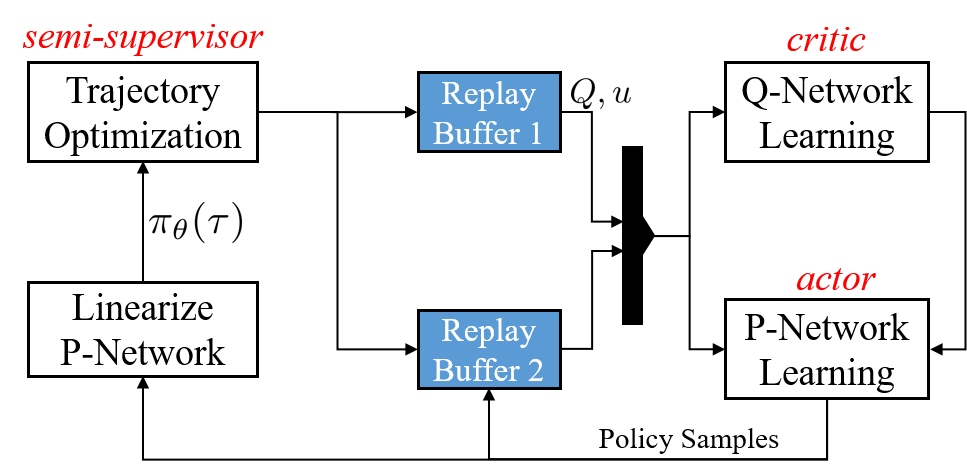}}
		\caption{Illustration of the proposed learning framework (guided-DDPG). Trajectory optimization provides initial guidance to both actor and critic nets to avoid excessive exploration. The actor-critic nets gradually establish the evaluation system, instead of relying on pure supervised learning.}
		\label{fig:guided-DDPG}
	\end{center}
\end{figure}

Figure~\ref{fig:guided-DDPG} illustrates the proposed guided-DDPG algorithm. 
Due to the discontinuity of the fitted dynamics in rigid precise systems, the trajectory optimization can have inconsistent behavior or requires dedicated parameter tuning. 
Therefore, a pure supervised learning from trajectory optimization cannot fulfill the task consistently. 
The actor-critic is incorporated to the framework to address this issue. The trajectory optimization serves as a \textit{semi-supervisor} to train the actor-critic to establish the initial critic and constrain the network in narrow task space. 
The involvement of the supervision will be reduced as the training progresses and the critic network becomes more accurate, since the actor-critic exhibits superior performance than the semi-supervisor. 

To be more specific, the trajectory optimization (semi-supervisor) has the following form: 
\begin{equation} 
\label{eq:trajopt_klcon}
\begin{aligned}
\min_{p} E_p(l(\tau)), \quad s.t. \ \  D_{KL}(p(\tau)\|\hat{p}_\theta(\tau)) \leq \epsilon,
\end{aligned}
\end{equation}
where $\hat{p}_\theta$ is set as the trajectory distribution generated by actor policy at the first sub-iteration, and is set as the previous trajectory distribution $\hat{p}$ for the successive $N_{trajopt}-1$ sub-iterations. 
Equation~(\ref{eq:trajopt_klcon}) is optimized by the dual:
\begin{equation} 
\label{eq:trajopt_lagrangian}
\begin{aligned}
\max_\eta \{\min_{p} E_p(l(\tau)) + \eta (D_{KL}(p(\tau)\|\hat{p}_\theta(\tau)) - \epsilon)\}.
\end{aligned}
\end{equation}

The optimization of $p$ is solved by LQG with fixed $\eta$ and dynamics, and the optimization of $\eta$ is done heuristically: decrease $\eta$ if $D_{KL}(p(
\tau)\| \hat{p}_\theta(\tau)) < \epsilon$, otherwise increase $\eta$. The trust region $\epsilon$ varies based on the expected improvement and actual one. $\epsilon$ would be reduced once the actual improvement is far smaller from the expected one, thus the network focuses on penalizing the KL divergence from $\hat{p}_\theta(\tau)$. 

We collect the trajectory after $N_{trajopt}$ sub-iterations to replay buffer $R_1$ for supervised training of actor-critic nets, and feed all the sample data during $N_{trajopt}$ executions to replay buffer $R_2$.  With the supervision from $R_1$, the critic is trained by: 
\begin{equation} 
\label{eq:guided_ddpg_critic}
\begin{aligned}
 & \phi \leftarrow \argmin_\phi\frac{1}{N_{dd}}\sum_{j = 1}^{N_{dd}}\left(y_j - {Q}_{{\phi}}(s_j, a_j)\right)^2 + \\ & \qquad \quad w_{to} \frac{1}{N_{to}}\sum_{i=1}^{N_{to}} \|{Q}_\phi(s_i, a_i) - Q_i^{to}\|^2\\
\end{aligned}
\end{equation} 
where $w_{to}, N_{to}$ are the weight and batch size of the semi-supervisor, $y_j$ has the same form as~(\ref{eq:ddpg_critic}). $(s_i, a_i, Q_i^{to})$ is the supervision data from $R_1$, and $(s_j, a_j, r_j, s_{j+1})$ is the sample data from $R_2$. 

The actor is trained by: 
\begin{equation} 
\label{eq:guided_ddpg_actor}
\begin{aligned}
\theta \leftarrow & \argmax_\theta\frac{1}{N_{dd}} \sum_{j = 1}^{N_{dd}} {Q}_{\hat{\phi}}(s_j, u_\theta(s_j)) + \\ & \quad w_{to} \frac{1}{N_{to}}\sum_{i=1}^{N_{to}} \|u_\theta(s_i) - a_i\|^2\\
\end{aligned}
\end{equation}

The supervision weight $w_{to}$ decays as the number of training rollouts $N_{roll}$ increases. We use $w_{to} = \frac{c}{N_{roll} + c}$, where $c$ is a constant to control the decay speed. 

The guided-DDPG algorithm is summarized in Alg.~\ref{alg:guided_ddpg}. The critic and actor are initialized in Line~\ref{alg:init}. Guided-DDPG runs for $EP$ epochs in total. In each epoch, semi-supervisor is first executed to update the trajectories for supervision. With the high stiffness, small clearance and the force/torque feedback, the fitted dynamics (Line~\ref{alg:fit_dynamics}) is discontinuous and has small trust region. Therefore, the trajectories generated from semi-supervisor might be sub-optimal. Nevertheless, they are sufficient to guide the initial training of the actor-critic. 
The actor-critic is trained in Line~(\ref{alg:ddpg} - \ref{alg:end_ddpg}) following the standard procedure of DDPG with the modified objective function (Line~(\ref{alg:ddpg_modifications})). 
The supervision weight $w_{to}$ is decreased as the training progresses due to the superior performance of the actor-critic than the semi-supervisor. 

\begin{algorithm} [t]
	\caption{Guided-DDPG}\label{alg:guided_ddpg}
	\begin{algorithmic}[1]
		\State \textbf{input:}$EP,N_{ddpg},N_{inc}, N_{trajopt},N_{roll}=0,R_{1/2}\leftarrow\Phi$\label{alg:input}
		\State \textbf{init:} $Q_\phi(s,a), u_\theta(s)$, set target nets $\hat{\phi}\leftarrow\phi, \hat{\theta}\leftarrow\theta$ \label{alg:init}
		\For {$epoch=0:EP$} \label{alg:trajopt}
		\State $p_{prev}\leftarrow u_\theta$
        \For {$it=0:N_{trajopt}$}
        \State $\mathcal{S}\leftarrow sample\_data(p_{prev})$, $R_2\leftarrow R_2\cup\mathcal{S}$
        \State $f_{dy}\leftarrow fit\_dynamics(\mathcal{S})$\label{alg:fit_dynamics}
        \State $\hat{p}_\theta \leftarrow linearize\_policy(p_{prev},\mathcal{S})$ 
        \State $p\leftarrow update\_trajectory(f_{dy},\hat{p}_\theta), p_{prev}\leftarrow p$
		\EndFor 
		\State $S\leftarrow sample\_data(p)$, $R_1\leftarrow R_1\cup\mathcal{S}, R_2\leftarrow R_2\cup\mathcal{S}$
		\For {$it=0:N_{ddpg}$} \label{alg:ddpg}
		\State $\mathcal{N}_{ex}\leftarrow exploration\_noise()$
		\State $s_0\leftarrow observe\_state(), w_{to} = \frac{c}{c+N_{roll}\texttt{++}}$ 
		\For {$t=0:T$}
		\State $a_t = u_\theta(s_t)+\mathcal{N}_{ex}(t)$, observe $s_{t+1},r_t$
		\State $R_2\leftarrow R_2\cup(s_t,a_t,s_{t+1},r_t)$
		\State sample $N_{to}, N_{dd}$ transitions from $R_1, R_2$
		\State update critic and actor nets by (\ref{eq:guided_ddpg_critic}) and (\ref{eq:guided_ddpg_actor})\label{alg:ddpg_modifications}
		\State update target nets by (\ref{eq:ddpg_target_update})
		\EndFor 
		\EndFor\label{alg:end_ddpg} 
		\State $N_{ddpg} \leftarrow N_{ddpg} + N_{inc}$
		\EndFor
	\end{algorithmic}
\end{algorithm}


\section{Simulations and Experiments}
\label{sec:experiments}
This section presents both the simulation and experimental results of the guided-DDPG to verify the effectiveness of the proposed learning framework. The videos are available at~\cite{website}.

\begin{figure}[t]
	\begin{center}
		{\includegraphics[width =0.9\linewidth]{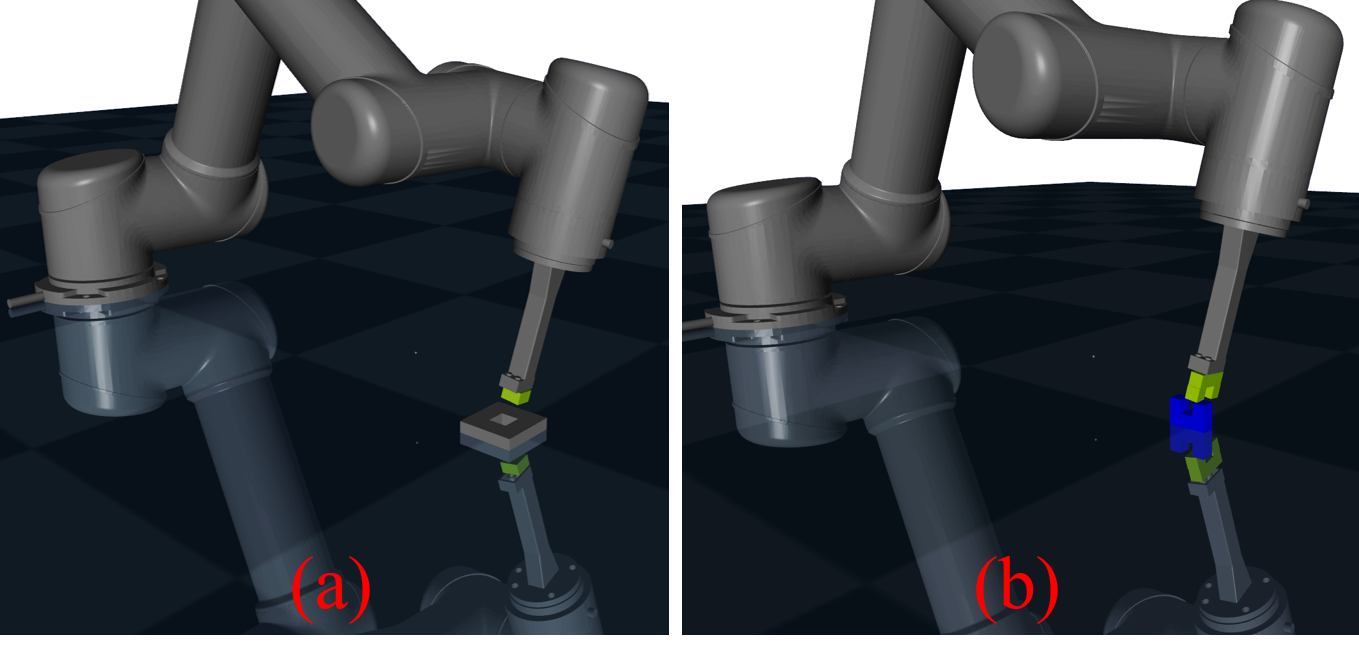}}
		\caption{Two simulation tasks for algorithm evaluation. (a) Lego brick insertion, (b) U-shape joint assembly. }
		\label{fig:simulation_environment}
	\end{center}
\end{figure}
\begin{figure*}[t]
	\begin{center}
		{\includegraphics[width =\linewidth]{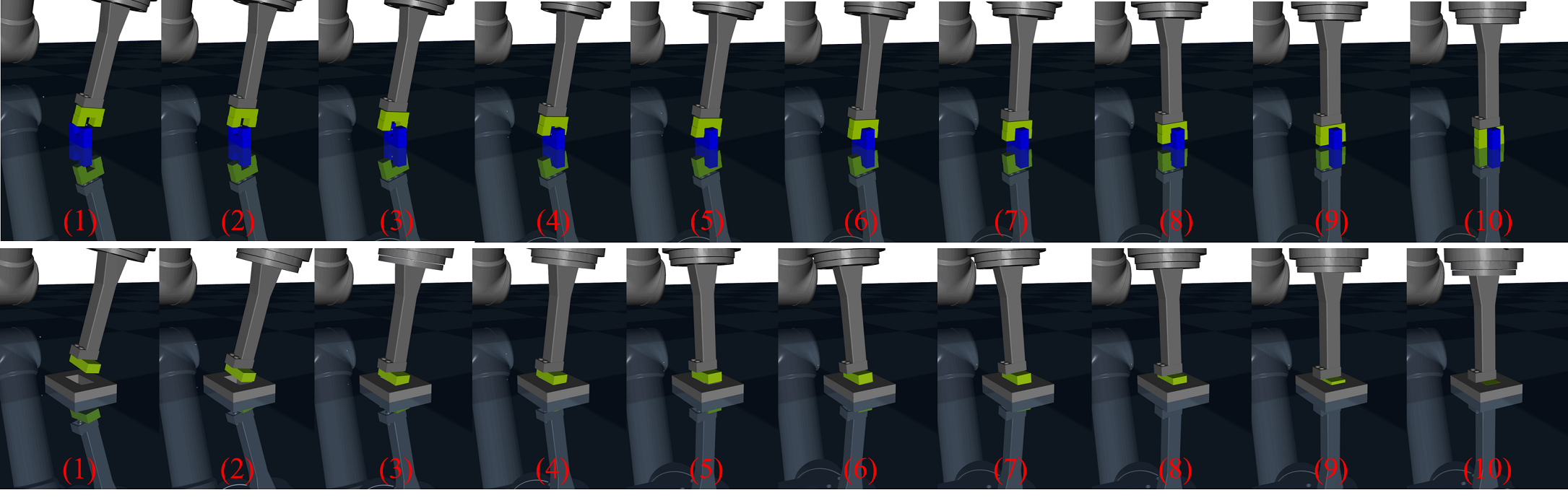}}
		\caption{Simulation animations of the proposed guided-DDPG on (Top) U-shape joint assembly and (Bottom) Lego brick insertion. The guided-DDPG was trained on 2$\times$2 Lego and tested on 4$\times$2 one. Snapshots are taken from left to right.}
		\label{fig:Usequence}
	\end{center}
\end{figure*}

To compare the performance of the guided-DDPG with other state-of-the-art RL algorithms, we built up a simulation model using the Mujoco physics engine~\cite{todorov2012mujoco}. The host computer we used was a desktop with 32GB RAM, 4.0GHz CPU and GTX 1070 GPU. 
A 6-axis UR5 robot model from universal robotics was used to perform the tasks. Two different assembly tasks were simulated, the first one was the Lego brick insertion, and the second one was the U-shape joint assembly, as shown in Fig.~\ref{fig:simulation_environment}. 

\subsection{Parameter Lists}
The number of the maximum epoch is set to $EP=100$, initial number of rollouts for DDPG and trajectory optimization were $N_{ddpg}=21$ and $N_{trajopt}=3$, respectively. To ensure less visit of trajectory optimization as the training progresses, we increased the number of rollouts by $N_{inc}=15$ for each DDPG iteration. The sizes of the replay buffer $R_1, R_2$ were 2000 and 1E6, respectively. 
The soft update rate $\gamma=0.001$ in (\ref{eq:ddpg_target_update}). The batch size for trajectory optimization $N_{to}$ and DDPG $N_{dd}$ were both 64. 
The algorithm used a cost function $l(s,a) = 0.0001\|a\|_2 + \|FK(s) - p_{tgt}(s)\|_2$, where $FK$ represents the forward kinematics and $p_{tgt}$ is the target end-effector points.

\subsection{Simulation Results}
\begin{figure}[t]
	\begin{center}
		{\includegraphics[width =0.9\linewidth]{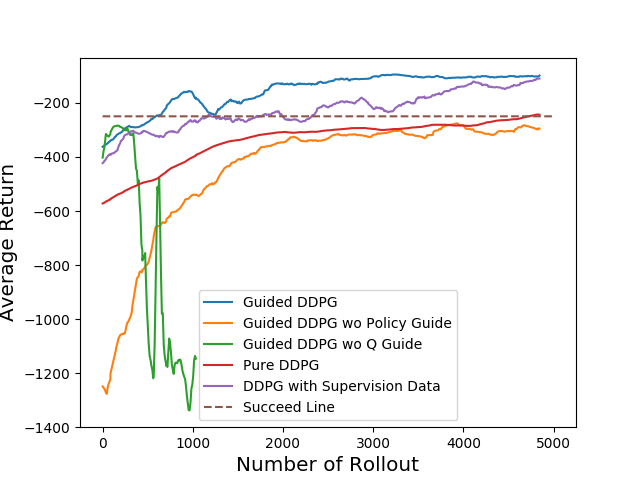}}
		\caption{Comparison of different supervisions with Lego brick insertion task. The supervision methods with performance in descending order: guided-DDPG (proposed), DDPG with supervised data in Replay buffer, pure DDPG, guided-DDPG w/o policy guidance, and guided-DDPG w/o critic guidance. }
		\label{fig:score_different_guidance}
	\end{center}
\end{figure}
\begin{figure*}[t]
	\centering
	\begin{subfigure}[b]{0.32\textwidth}
		\centering
		\includegraphics[width=\textwidth]{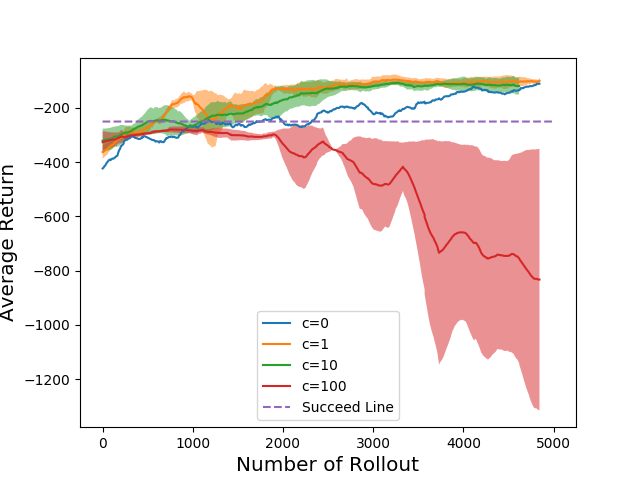}
		\caption{}
		\label{fig:score_different_weights}
	\end{subfigure}
	\hfill
	\begin{subfigure}[b]{0.32\textwidth}
		\centering
		\includegraphics[width=\textwidth]{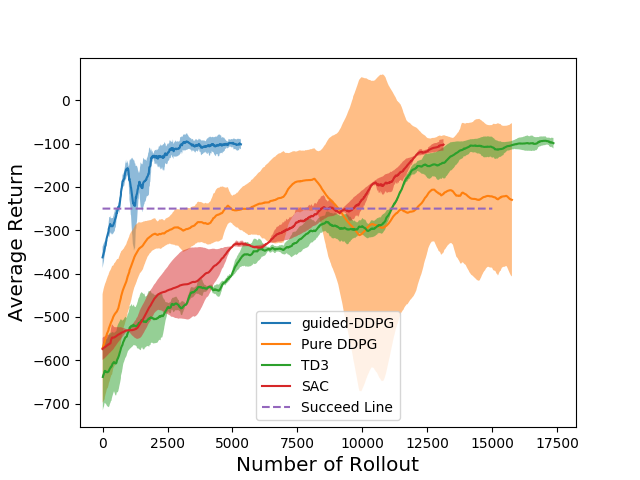}
		\caption{}
		\label{fig:score_different_methods}
	\end{subfigure}
	\hfill
	\begin{subfigure}[b]{0.32\textwidth}
		\centering
		\includegraphics[width=\textwidth]{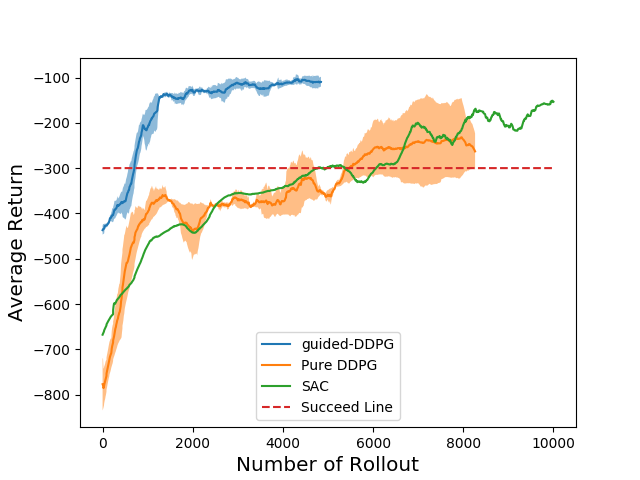}
		\caption{}
		\label{fig:score_different_U}
	\end{subfigure}
	\caption{(a) Illustration of the supervision weights on Lego brick insertion task. (b) Comparison of the algorithms for Lego brick insertion task. (c) Comparison of the algorithms for U-shape joint assembly task.}
	\label{fig:three graphs}
\end{figure*}

The simulation results on U-shape joint assembly and Lego brick insertion are shown by Fig.~\ref{fig:Usequence}. Both  simulations were trained with assembly clearance as 0.1 mm. Guided-DDPG takes poses and force/torque measurements of the end-effector as the states, and generates joint torques as action to drive the robot. 
The U-shape joint has more complicated surface than the Lego brick, and a successful assembly requires matching the shapes twice, as shown in Fig.~\ref{fig:Usequence} (Top). Despite the difficulties, the proposed algorithm was able to train the policy within 1000 rollouts. We also visualized the adaptability of the trained policy on the Lego brick insertion task, as shown in Fig.~\ref{fig:Usequence} (Bottom). The policy was trained with a brick of size 2$\times$2 and clearance 0.1 mm and tested with a brick of size 4$\times$2 and clearance 1 $\upmu$m. Moreover, the brick position had an unknown offset (1.4 mm) to the network. The proposed network was able to address these uncertainties and successfully inserted the brick to a tighter hole with uncertain position.

\subsubsection{Comparison of Different Supervision Methods}
The proposed learning framework guides both the critic and actor. To illustrate the necessity of the proposed guidance, we compared the results of guided-DDPG with several other supervision methods, including the guided-DDPG with partial guidance, pure-DDPG with supervision data to replay buffer (no supervision on objective function) and the pure-DDPG. The result was shown in Fig.~\ref{fig:score_different_guidance}. The proposed guided-DDPG achieved the best performance. 
The partial guidance without critic  (Fig.~\ref{fig:score_different_guidance} Green) was able to guide the actor and realized safe exploration at the beginning. However, the actor network behaved worse as the involvement of the semi-supervisor reduced and the weight of the critic increased, since the critic is trained purely by the contaminated target actor~(\ref{eq:ddpg_critic}). In contrast, the partial guidance without actor~(Fig.~\ref{fig:score_different_guidance} Orange) had poorly behaved actor since the actor was trained purely by the policy gradient using the contaminated critic~(\ref{eq:ddpg_actor}). 
The pure-DDPG with supervision data~(Fig.~\ref{fig:score_different_guidance} Purple) achieved better performance than pure-DDPG, since the trajectories obtained from semi-supervisor were better behaved than the initial rollouts of DDPG. This kind of supervision is similar with the human demonstration in~\cite{vecerik2017leveraging}.

\subsubsection{Effects of the Supervision Weight $w_{to}$}
The supervision weight $w_{to}$ balances the model-based supervision and model-free policy gradient in actor/critic updates, as shown in (\ref{eq:guided_ddpg_actor}) and (\ref{eq:guided_ddpg_critic}). The results of different weights on Lego brick insertion are shown in Fig.~\ref{fig:three graphs} (a). With $c = 1$, the supervision weight is $w_{to} = \frac{1}{1+N_{roll}}$.  The weights starts with 1 and decays to 0.001 as $N_{roll}=1000$, while $c=100$ makes $w_{to}$ decay to 0.1 as $N_{roll}= 1000$. Slower decay provides excessive guidance by the semi-supervisor and contaminates the original policy gradient and makes the DDPG unstable. Empirically, $c=1\sim10$ achieves comparable results. 

\subsubsection{Comparison of Different Algorithms}
The proposed learning framework was compared with other state-of-the-art algorithms, including the pure-DDPG, twin delayed deep deterministic policy gradients (TD3)~\cite{td3} and the soft actor-critic (SAC)~\cite{sac}. Default parameters were used for TD3, as shown in~\cite{rlkit}. As for SAC, we used the default parameters in~\cite{rlkit} with tuned reward scale as 10. The comparison result on Lego brick insertion task is shown in Fig.~\ref{fig:three graphs} (b). The proposed guided-DDPG passed the success threshold (shaded purple line) at the 800 rollouts and consistently succeeded the task after 2000 rollouts. In comparison, the pure DDPG passed the success threshold at the 5000 rollouts and collapsed around 10000 rollouts. The performance of pure DDPG was inconsistent in seven different trials. TD3 and SAC had the similar efficiency with pure DDPG. The comparison of the algorithms on U-shape joint assembly is shown in Fig.~\ref{fig:three graphs} (c).  Similar with Lego brick insertion, the guided-DDPG achieved more stable and efficient learning. The time efficiency and data-efficiency of the DDPG and guided-DDPG are compared in Table~\ref{tab:comparison}. 
\begin{table}[t]
	\centering
	\caption{Comparison between DDPG and guided-DDPG}
	\label{tab:comparison}
\begin{tabular}{c|c|c}
  items  &   DDPG & Guided-DDPG\\
\hline 
time (min)         & 83  & \textbf{37.3} \\
\hline
data (rollouts)          & 7000  & \textbf{1500}  \\
\hline 
\end{tabular}
\end{table}
\begin{figure*}[ht]
	\centering
	\begin{subfigure}[b]{0.214\textwidth}
		\centering
		\includegraphics[width=0.9\textwidth]{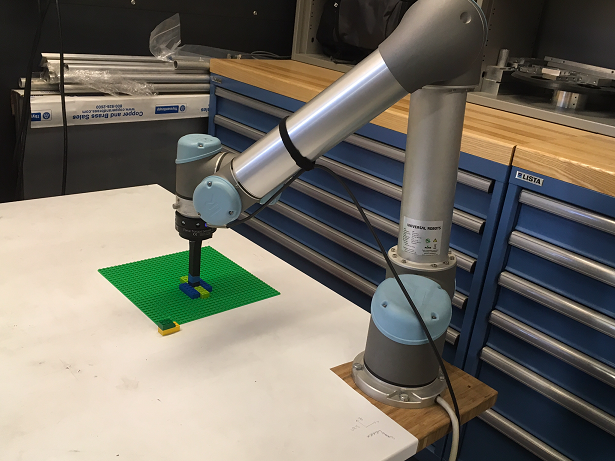}
		\caption{}
		\label{fig:exp_setup}
	\end{subfigure}
	\hfill
	\begin{subfigure}[b]{0.77\textwidth}
		\centering
		\includegraphics[width=0.9\textwidth]{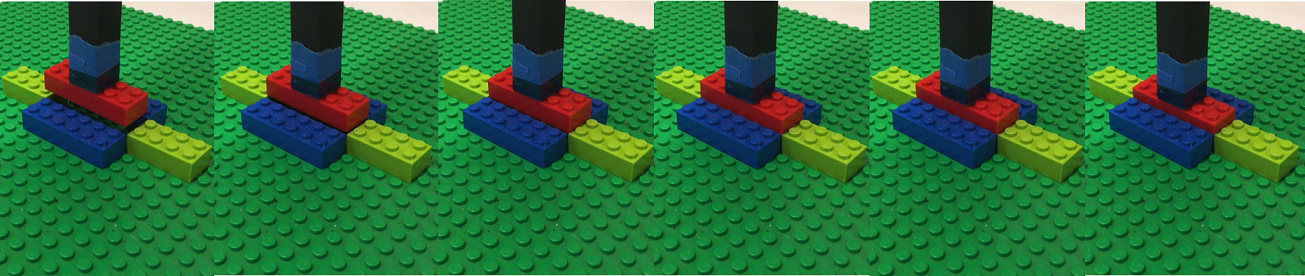}
		\caption{}
		\label{fig:exp_results}
	\end{subfigure}
	\caption{(a) Experimental setup, and (b) experimental results for Lego brick insertion.}
	\label{fig:two_graphs}
\end{figure*}
\subsubsection{Adaptability of the Learned Policy}
\begin{figure}[t]
	\begin{center}
		{\includegraphics[width =0.9\linewidth]{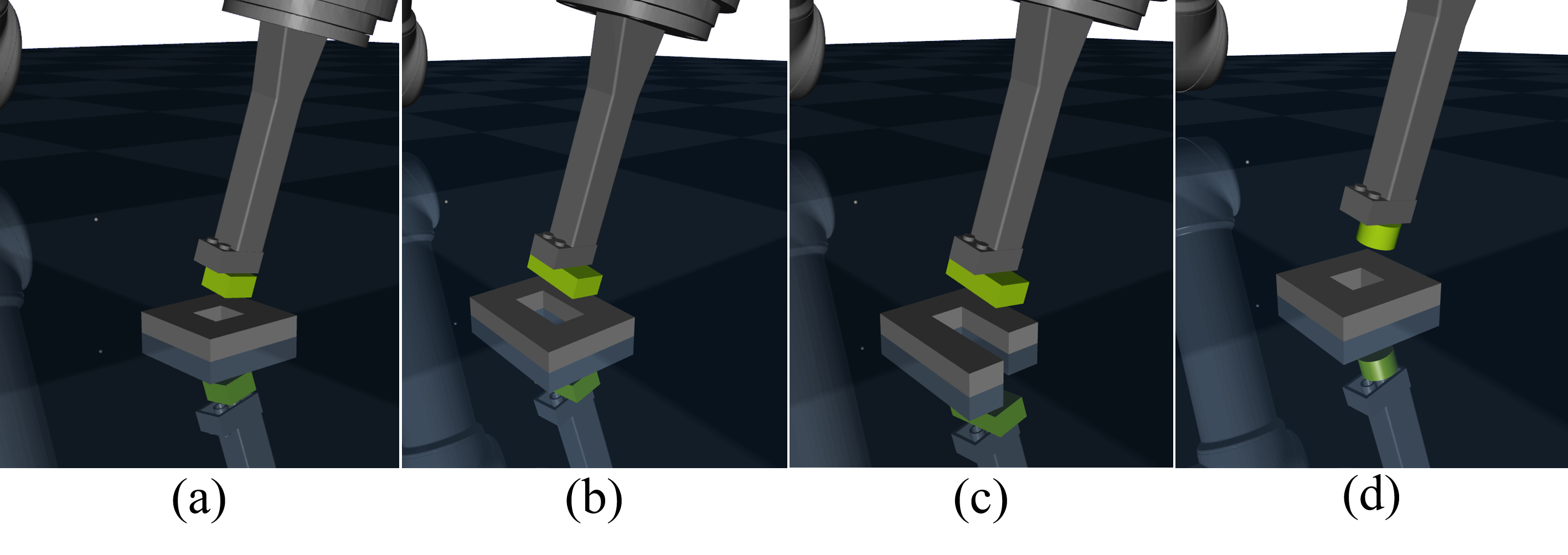}}
		\caption{Different shapes of the bricks and holes for adaptability test. (a) 2$\times$2 brick used in training, (b) 4$\times$2 brick, (c) 4$\times$2 brick with incomplete hole, and (d) cylinder brick. }
		\label{fig:adaptability}
	\end{center}
\end{figure} 
The adaptability of the learned policy is discussed in this section. Three different types of uncertainties were considered. The first type was the unknown hole position. The learned policy was able to successfully insert the brick when moving the hole to an uncalibrated position (maximum offset is 5 mm, hole has width of 16 mm). The second type of uncertainty was the shapes of peg/hole. We found that the learned policy is robust to different shapes shown in Fig.~\ref{fig:adaptability}. The third type was the different clearance. The policy was trained with clearance 0.1 mm and tested successfully on insertion tasks with clearance 10 $\upmu$m, 1 $\upmu$ and 0. 
The simulation videos are available at~\cite{website}. 
%

\subsection{Experimental Results}
Experimental results are presented in this section. The Lego brick was attached to a 3D printed stick at the end-effector of the Universal robot (UR5). A Robotiq FT 300 force torque sensor was used to collect the force/torque signal at the wrist. The experimental setup is shown in Fig.~\ref{fig:two_graphs}(a). 
The policy took the estimated hole position and the force/torque reading as inputs, and generated transitional velocities for the end-effector. The velocity was tracked by a low-level tracking controller. 
The clearance of the Lego brick is less than 0.2 mm. The target position of the hole had 0.5 mm uncertainty, yet the policy was able to successfully locate the hole and insert the brick, as shown in Fig.~\ref{fig:two_graphs}(b). 
It took 2 hours for pure-DDPG to find a policy in the exploration space bounded within $1$ mm around the hole, and took 1.5 hours for guided-DDPG to find a policy in a larger exploration space bounded within $3$ mm around the hole. 
The experimental videos are shown in~\cite{website}.

\section{Conclusions and Future Works} 
\label{sec:conclusions}
This paper proposed a learning framework for high precision assembly task. The framework contains a trajectory optimization and an actor-critic structure. The trajectory optimization was served as a semi-supervisor to provide initial guidance to actor-critic, and the critic network established the ground-truth quality of the policy by learning from both the semi-supervisor and exploring with policy gradient. The actor network learned from both the supervision of the semi-supervisor and the policy gradient of the critic. The involvement of critic network successfully addressed the stability issue of the trajectory optimization caused by the high-stiffness and the force/torque feedback. 
The proposed learning framework constrained the exploration in a safe narrow space, improved the consistency and reliability of the model-based RL, and reduced the data requirements to train a policy. Simulation and experimental results verified the effectiveness of the proposed learning framework. 

In the future, the authors would evaluate the algorithm on more realistic industrial applications such as connector insertion, furniture assembly and tight peg-in-hole tasks.

\addtolength{\textheight}{-1cm}   



 
\section*{Acknowledgment}
The authors would like to thank Dr. Yotto Koga and AI Lab in Autodesk Inc. for the help on experiments. 

\bibliographystyle{IEEEtran}
\bibliography{references}

\end{document}